\documentclass{article}
\usepackage{spconf,amsmath,graphicx}
\usepackage{comment}
\usepackage{booktabs}

\title{Distilling Facial Knowledge With Teacher-Tasks:\\
Semantic-Segmentation-Features for Pose-Invariant Face-Recognition}
\name{Ali Hassani, Zaid El Shair, Rafi Ud Duala Refat, Hafiz Malik \thanks{Thanks to Ford Motor Company for Alliance Grant Biometric Forensics. © 2022 IEEE. Personal use of this material is permitted. Permission from IEEE must be obtained for all other uses, in any current or future media, including reprinting/republishing this material for advertising or promotional purposes, creating new collective works, for resale or redistribution to servers or lists, or reuse of any copyrighted component of this work in other works.}}
\address{Department of Electrical and Computer Engineering, \\
University of Michigan - Dearborn, Dearborn, USA}
\begin{document}
%
\maketitle
\begin{abstract}
This paper demonstrates a novel approach to improve face-recognition pose-invariance using semantic-segmentation features. The proposed Seg-Distilled-ID network jointly learns identification and semantic-segmentation tasks, where the segmentation task is then ``distilled'' (MobileNet encoder). Performance is benchmarked against three state-of-the-art encoders on a publicly available data-set emphasizing head-pose variations. Experimental evaluations show the Seg-Distilled-ID network shows notable robustness benefits, achieving 99.9\% test-accuracy in comparison to 81.6\% on ResNet-101, 96.1\% on VGG-19 and 96.3\% on InceptionV3. This is achieved using approximately one-tenth of the top encoder's inference parameters. These results demonstrate distilling semantic-segmentation features can efficiently address face-recognition pose-invariance.
\end{abstract}

\section{Introduction}\label{sec:intro}
Face-recognition (FR) is becoming the go-to authentication technology for access control and verification applications. Its popularity starts with evolution of smart phones, where over 100 million devices offer it as seamless-unlock method \cite{pascu2020phones}. This has led other industries to follow suit, where commercial real-estate \cite{lwin2015automatic}, aviation \cite{reservations_airport} and banking \cite{heun2021banks} now use FR as a means to differentiate customer experience. This is made possible by advances in deep-learning \cite{schroff2015facenet}; state-of-the-art models can now discern 1 cooperative face from over 50,000 \cite{androidbiospec}. Having robust tolerance to pose-variations, however, is still a challenge \cite{zhang2009pose}.

Pose-variations are facial rotations over yaw and pitch. These change the relative-position of key-points (e.g., nose, eyes) and introduce variance within identity classes. As such, FR algorithms can struggle to discern the same person rotating from different people \cite{zhang2009pose}. In particular when applying stringent industry false-acceptance-rate thresholds \cite{androidbiospec}, variations in pose often result in false-rejections \cite{zhang2009pose}.

Current state-of-the-art methods rely on alignment techniques and/or sophisticated loss-functions to address pose-variability. Through alignment pre-processing, algorithms can project a cooperative-face for identification \cite{jin2017alignment}. Alternatively, contrastive loss-functions (such as triplet) implicitly address pose-variations through relative class-distance \cite{schroff2015facenet}. While notable advances, even best-in-class algorithms struggle to achieve 100\% accuracy on competition data-sets \cite{huang2008lfw}. 

\begin{figure}[t]
 \centerline{
 \includegraphics[width = .5\textwidth]{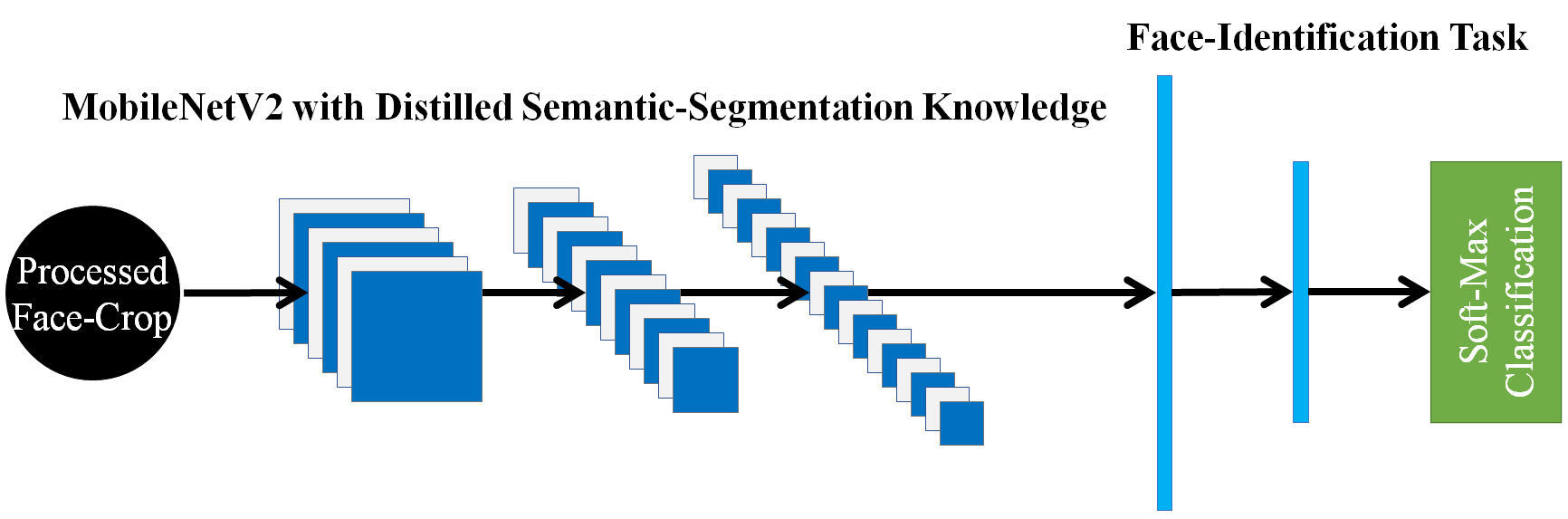}}
 \caption[Seg-Distilled-ID Network for Pose-Invariance.]{Seg-Distilled-ID Network for Pose-Invariance.}
 \label{fig:BJL-distilled-network}
\end{figure}

This paper presents the Seg-Distilled-ID network. This is a new approach to knowledge-distillation, using a teacher-task in lieu of a teacher-network. The Seg-Distilled-ID network is first jointly trained on both identification and (teacher) semantic-segmentation tasks, where the teacher-task is then removed. This ``distills'' the semantic-structures as context for precise identification (see Fig. \ref{fig:BJL-distilled-network}). Recognition accuracy is benchmarked against three state-of-the-art encoders
on the Mut1ny commercial face-segmentation data-set \cite{mut1ny2021faceseg} (11,830 images selected from 67 subjects, varied over pose and lighting). The proposed Seg-Distilled-ID network achieves best-in-class accuracy using 2.4M inference parameters. These results demonstrate distilling semantic-segmentation features can efficiently address face-recognition pose-invariance.

In summary, this paper makes the following contributions:
\begin{itemize}
\item Novel knowledge-distillation method via teacher-tasks.
\item Best-in-class ID accuracy with efficient parameter space (99.9\%,  Mut1ny faces \cite{mut1ny2021faceseg}).
\end{itemize}

\begin{figure*}[ht]
 \centerline{
 \includegraphics[width = .85\textwidth]{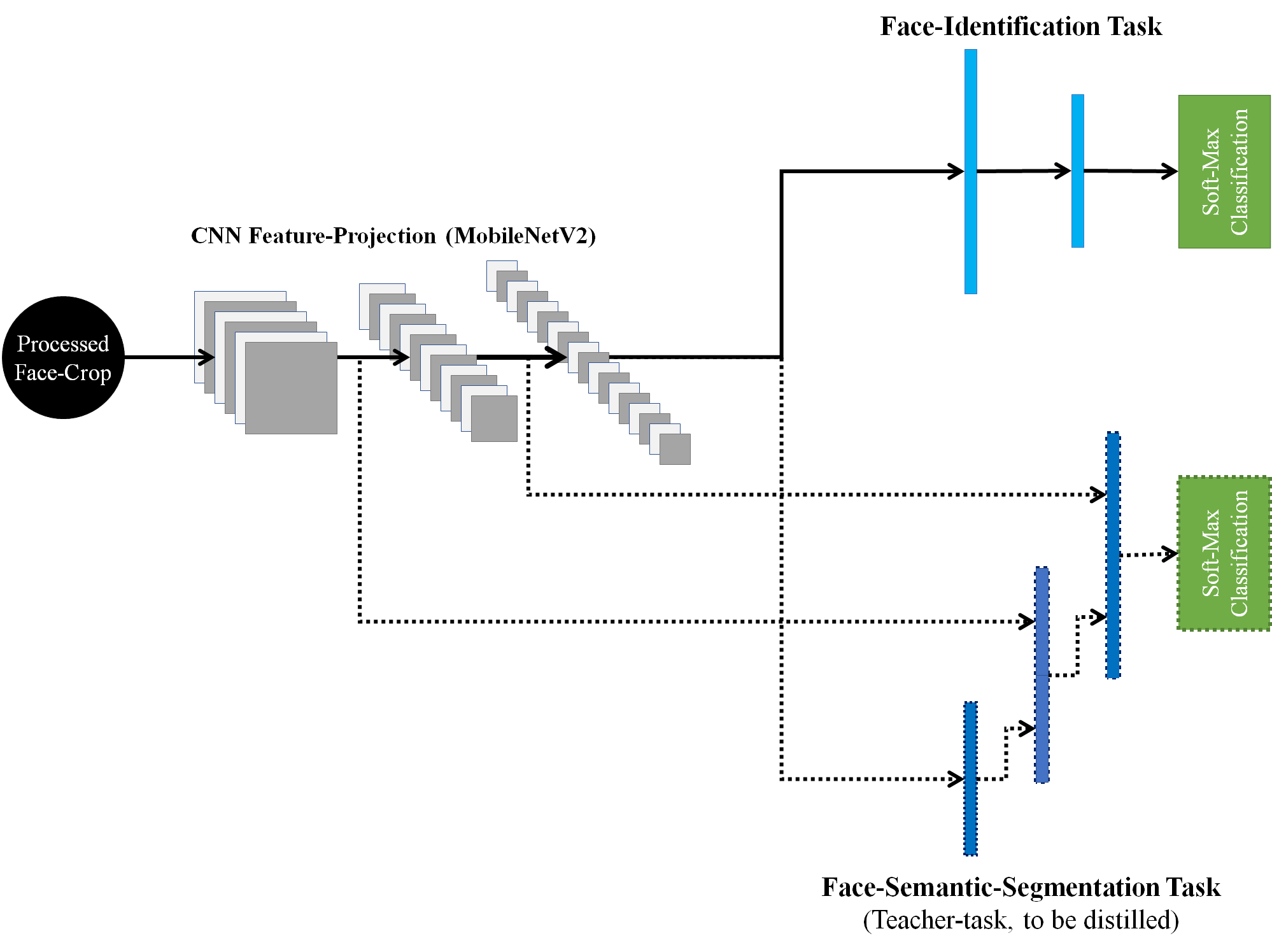}}
 \caption[Improving Face-Identification Context with Semantic-Segmentation Teacher.]{Improving Face-Identification Context with Semantic-Segmentation Teacher.}
 \label{fig:BJL-seg-id-network}
\end{figure*}

\section{Related Works}\label{sec:works}
Face-recognition (FR) research starts in the 1970s as a template-matching problem. This is pioneered by the discovery that statistical-distributions (e.g., Eigenfaces \cite{turk1991eigenfaces}) are generally robust. This is expanded upon by using hand-crafted features to describe distinguishable features \cite{moghaddam1998beyond}. These features, however, are insufficient at large-scales. This ultimately transitions to deep-learning (DL) solutions (starting with DeepFace \cite{taigman2014deepface}). Today, state-of-the-art DL networks emphasize a combination pose-alignment pre-processing \cite{jin2017alignment} (which may include 3-D projection \cite{hu2017ldf, zhou2018gridface}) and contrastive loss-functions \cite{schroff2015facenet}.

Despite all these advances, pose-variations are still a challenge \cite{zhang2009pose}. To address this, a new trend is to apply multi-task-learning (MTL) for sharing context. This is first exemplified by Ranjan et al, who combine landmarks and pose tasks on a face-detection network (HyperFace) to improve reliability \cite{ranjan2017multi_hyperface}. Others have now recently applied this approach to identification. Yin and Xiaoming have a pose estimation task connected to the identification features \cite{yin2017multi_face_pose}. Wang et al alternatively use semantic descriptors of sub-structures: size of eyes, nose and cheeks \cite{wang2017multi_face_attribute}. Both approaches show consistent (small) improvements on FR competition data-sets. 

This research differentiates on these findings by using a knowledge-distillation approach with a precise-descriptor: semantic-segmentation. The aforementioned approaches share common features with the identification task. In this case, the network is jointly trained on facial-structure with identification, then ``distills'' the teacher-task. The ``distilled'' semantic-features enable the encoder to generate precise features, enabling efficient pose-invariance recognition.
\section{Methodology} \label{chap:bio_joint_learning-methods}
This research proposes a novel application of Multi-Task-Learning (MTL) to improve face-recognition pose-robustness. Facial semantic-segmentation is ``distilled'' by using a ``teacher-task.'' By encoding relative facial-structure, the loss function can better discern \textit{inter} versus \textit{intra} class variations.

\subsection{Seg-Distilled-ID Network}\label{sec:BJL-seg-id}
Fig. \ref{fig:BJL-seg-id-network} shows how the Seg-Distilled-ID network starts with identification and segmentation tasks. The segmentation-task functions as a teacher, helping the ID-task better converge towards optimal weights. Once training is complete the teacher-task is removed (note the dashed lines).

The network assumes a U-Net architecture \cite{ronneberger2015unet}. U-Net is selected both for its applications to biomedical semantic-segmentation \cite{ronneberger2015unet} and option for efficient MobileNetV2 encoder \cite{sandler2018mobilenetv2}. A MobileNetV2 backbone \cite{sandler2018mobilenetv2} encodes features for parallel identification and semantic-segmentation tasks. The identification-task is constructed by applying a global-average pooling layer, followed with a dense, 128-neuron, feature layer (ReLU activation \cite{agarap2018relu}) and a dense, 67-neuron, classification layer (soft-max activation \cite{liu2016softmax}). The segmentation-task is constructed using the Pix2Pix decoding layers \cite{isola2017pix2pix} (e.g. final segmentation output of 128 by 128).

Both tasks use a categorical-cross-entropy loss, as shown in \eqref{eq:BJL-cross-entropy}. This better separates out the (log) distance between classes by incorporating probability of the observation, $o$, belonging to the label-class, $c$. This probability can be defined as $p(o, c)$ \cite{liu2016softmax}. A binary label, $\hat y$, indicates whether the prediction matches the correct class. This is done per class $c$ of $M$ \cite{liu2016softmax} in an expected-value fashion.

\begin{equation}
    CE = -\sum_{c=1}^{M} \hat y_{o,c} \log{p(o, c)}
    \label{eq:BJL-cross-entropy}
\end{equation}

Equation \eqref{eq:BJL-joint-loss} shows the joint MTL-loss as a linear combination. Both identification and semantic-segmentation are multi-class-tasks, employing (categorical) cross-entropy loss. The losses are weighted in a 10 to 1 ratio; this is because the segmentation-task is both inherently harder and functions as the ``teacher'' for ``distillation.'' This is described in \eqref{eq:BJL-joint-loss}, where $CE$ is the cross-entropy loss function, $Y$ and $\hat{Y}$ are the respective task inference and label vectors, and $\lambda$ is the loss-weight (i.e., 1 and .1 respectively).

\begin{equation}
    Loss = \lambda_{Seg} \cdot CE(Y_{Seg},\hat{Y}_{Seg}) 
    + \lambda_{ID} \cdot CE(Y_{ID},\hat{Y}_{ID})
    \label{eq:BJL-joint-loss}
\end{equation}

Once training is complete, the teacher segmentation-task-layers are removed. This significantly reduces the network parameters, from 6.5M to 2.4M, for inference. Fig. \ref{fig:BJL-distilled-network} shows the final inference structure (see: first-page), where the encoder color change represents the segmentation knowledge-distillation. The purpose of this architecture is to both retain efficiency while demonstrating the dark-knowledge of the segmentation task is sufficient to improve identification.
\section{Performance Evaluation}\label{chap:bio_joint_learning-performance-evaluatioon}
This experiment evaluates the identification accuracy when introducing significant pose-variations. The purpose is to demonstrate the utility of distilling face-segmentation as contextual features. The Seg-Distilled-ID network is validated against identification networks using MobileNetV2 \cite{sandler2018mobilenetv2} without segmentation-context and three state-of-the-art network encoders.

\begin{figure}[h]
 \centerline{
 \includegraphics[width = .4\textwidth]{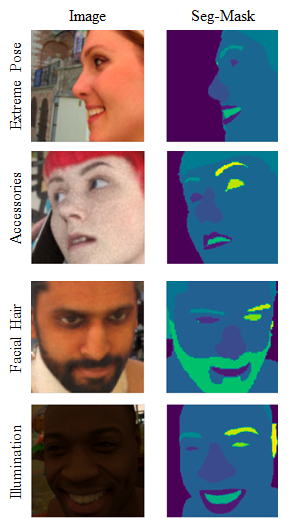}}
 \caption[Mut1ny data-set challenging image samples.]{Mut1ny data-set challenging image samples.}
 \label{fig:BJL-mut1ny}
\end{figure}

\subsection{Experiment: Pose-Invariant Identification}\label{sec:bio_joint_learning-exp-1}
This experiment evaluates identification performance under high pose-variation. The Mut1ny Face/Head Segmentation (commercial edition) data-set \cite{mut1ny2021faceseg} is used, employing 67 synthetic users with 150-250 unique perspectives (pose and background) each (11830 total). Each face is annotated with 14 structure classes: lips, left-eye, right-eye, nose, skin, hair, left-eyebrow, right-eyebrow, left-ear, right-ear, teeth, facial-hair, spectacles and background. These are cropped using the Dlib face detection tool \cite{king2009dlib}. Model verification-accuracy is measured following Labelled Faces in the Wild procedures \cite{huang2008lfw}. Each person has 90\% of their face-perspectives associated for training (8,320) and validation (2,080); test accuracy is evaluated on the remaining 10\% (1,430).

Fig. \ref{fig:BJL-mut1ny} shows some sample images (with segmentation-masks) from the evaluation data-set. While there are only 67 people, it is a very challenging face-recognition data-set. There are differences in pose, accessories, facial-hair and illumination. These significantly increase \textit{intra}-class variability.

\subsection{Benchmark Algorithms}
The Seg-Distilled-ID network is benchmarked against three state-of-the-art encoders and MobileNetV2 without teacher-task \cite{sandler2018mobilenetv2}. Note that this is a comparison of encoder knowledge where inputs and ID-loss-function are kept identical. Furthermore, a comparison of pose-estimation versus semantic-segmentation context is viewed as relevant due to the work of Yin \cite{yin2017multi_face_pose}. However, given the Mut1ny data-set does not contain the same pose-annotations, it is not pragmatic to do so. Evaluating input transformation, loss-function and task-sharing approaches are viewed as key next steps.

Each benchmark network follows the same ID task-structure. That is to say an encoder generates the features, where are global-average-pooled, then classified using a 128-neuron dense feature-layer (ReLU activation) \cite{agarap2018relu} and 67-neuron dense ID-classification-layer (soft-max activation) \cite{liu2016softmax}. The following network feature-encoders are used:
\begin{enumerate}
    \item MobileNetV2 \cite{sandler2018mobilenetv2}
    \item ResNet-101 \cite{he2016resnet}
    \item VGG-19 \cite{simonyan2014vgg}
    \item IncepvtionV3 \cite{szegedy2016inception}
\end{enumerate}

Each network is referred to as the encoder ``-ID''. E.g., validation-network 1 is designated ``MobileNetV2-ID.'' All feature-encoders come pre-trained on ImageNet \cite{deng2009imagenet}. Networks are compiled and trained in the same fashion, up to 125 epochs with a validation-loss patience of 20. Due to space constraints, training and validation curves are not shown. 
\section{Evaluation Results}
Table \ref{tab:face-rec-performance} shows the performance evaluation results. As generally expected, having a stronger encoder correlates with better ID classification. All networks but MobileNetV2-ID train to a validation accuracy of at least 95\% (training data not shown for space). This understandable from the encoder architectures. For example, InceptionV3-ID network has a relatively-high parameter-count with factorized-convolutions \cite{szegedy2016inception} and trains robustly. Conversely, the MobileNetV2-ID stops early and clearly over-fits due to its efficient design. 

\begin{table}[h]
\begin{center}
\caption{Network evaluation on Mut1ny data-set.}
\label{tab:face-rec-performance}
\begin{tabular}{l c c }
\toprule
\textbf{Network} & \textbf{Parameters} & \textbf{Test Accuracy}  \\
\midrule
                MobileNetV2-ID &	2.4M	& 21.9\% \\
                ResNet-101-ID  &	43M	& 81.6\% \\
                VGG-19-ID  &	20M	& 96.1\% \\
                InceptionV3-ID  &	22M	& 96.3\% \\
                \textbf{Seg-Distilled-ID}  &	\textbf{2.4M (6.5M +Seg)} & \textbf{99.9\%}  \\
                
\bottomrule
\end{tabular}
\end{center}
\end{table}

This performance disparity exemplifies the benefits of distilling semantic-segmentation features. Despite MobileNetV2-ID over-fitting, the Seg-Distilled-ID has the highest accuracy score evaluated. This is achieved while retaining the MobileNet architecture's efficiency (approximately one-tenth of the VGG and Inception network parameters). The parenthesis indicates that 2.4M parameters are used for inference and 6.5M are used for jointly training with the teacher-task. 

The parameter efficiency is explainable by using the semantic-segmentation knowledge to select optimal features. Top-tier encoders use large parameter-spaces to implicitly infer context, enabling them to perceive information the base MobileNetV2-ID cannot. This methodology instead explicitly provides context through the facial-structure teacher-task. Fig. \ref{fig:BJL-mut1ny} shows the semantic-segmentation masks; one can infer how the features that encode facial-structure variation across pose enable precise identification across pose. This feature robustness enables the Seg-Distilled-ID to efficiently achieve best-in-class performance.

Note that generalized pose robustness is very much novel. Others demonstrate re-aligning the face in 3-D space can improve identification robustness (such as LDF-Net \cite{hu2017ldf} and GridFace \cite{zhou2018gridface}). These methods are effective but degrade as yaw and pitch increase. It is hypothesized the 3-D alignment algorithms synthetically inferring the obscured facial features cascades bias from the projector. The Seg-Distilled-ID avoids this bias by learning facial-structures in a one-shot approach.
\section{Conclusions}\label{chap:bio_joint_learning-conclusion}
This paper presents the Seg-Distilled-ID network to address pose-invariance face-recognition. This is a novel application of knowledge-distillation, where an ID-task is jointly-trained with a ``distilled'' teacher semantic-segmentation-task. Benchmarking with state-of-the-art encoders ResNet-101 \cite{he2016resnet}, VGG-19 \cite{simonyan2014vgg} and InceptionV3 \cite{szegedy2016inception} shows the proposed Seg-Distilled-ID network achieves best-in-class performance using minimal parameters (MobileNetV2 \cite{sandler2018mobilenetv2} encoder).

Next steps include larger-scale evaluation with varied context-encoding methods. The Mut1ny data-set \cite{mut1ny2021faceseg} has only 67 subjects in the synthetic-face repository at this time; hence, the planned next step is attempt transfer-learning these features onto Labelled Faces in the Wild \cite{huang2008lfw}. To accommodate the increase in ID classes, a comparison of U-Net \cite{ronneberger2015unet} and DeepLabV3 \cite{chen2017deeplab} designs will be done with various task-architectures. Benchmarking will also include face-alignment pre-processing and contrastive loss-functions.
\section{Acknowledgments}
The authors would like to give special thanks to Ford Motor Company for funding this research via University Alliance Grant. This in particular includes Ford engineers Justin Miller and Jon Diedrich for their continued support.
\bibliographystyle{IEEEbib}
{\small
\bibliography{References/main}}

\end{document}